\documentclass{article}

\usepackage[preprint]{neurips_2026}

\usepackage[utf8]{inputenc}
\usepackage[T1]{fontenc}
\usepackage{hyperref}
\usepackage{url}
\usepackage{booktabs}
\usepackage{amsfonts}
\usepackage{amsmath}
\usepackage{nicefrac}
\usepackage{microtype}
\usepackage{xcolor}
\usepackage{multirow}
\usepackage{graphicx}
\usepackage{xspace}
\usepackage{colortbl}
\usepackage{arydshln}
\definecolor{cellblue}{HTML}{9DC3E6}
\definecolor{hdrblue}{HTML}{2E5FA3}

\newcommand{\avg}{\textsc{Avg}}

\newcommand{\opsd}{\textsc{Opsd}}
\providecommand{\method}{S$^2$VOPD\xspace}
\newcommand{\best}[1]{\textbf{#1}}
\newcommand{\second}[1]{\underline{#1}}

\title{Self-Supervised Visual On-Policy Distillation}

\author{
Yijiang Li$^{\spadesuit}$ \quad
Yijun Liang$^{\diamondsuit}$ \quad
Yunjie Tian$^{\clubsuit}$ \quad
Bingyang Wang$^{\heartsuit}$ \quad
\\
\textbf{Ke Zhang}$^{\dagger}$ \quad
\textbf{Zhenfei Yin}$^{\star}$ \quad
\textbf{Di Fu}$^{\clubsuit}$ \quad
\textbf{Philip Torr}$^{\star}$ \quad
\textbf{Nuno Vasconcelos}$^{\spadesuit}$ \\[0.35em]
$^{\spadesuit}$UC San Diego \quad
$^{\diamondsuit}$University of Maryland, College Park \quad
$^{\heartsuit}$Georgia Institute of Technology \quad
\\
$^{\dagger}$Johns Hopkins University \quad
$^{\star}$University of Oxford \quad
$^{\clubsuit}$Independent Researcher \quad
\\[0.3em]
\texttt{\{yijiangli, nuno\}@ucsd.edu}
}

\begin{document}

\maketitle

\begin{abstract}

Visual on-policy distillation relies heavily on an informative teacher--student asymmetry, through either a larger, stronger teacher or privileged supervision, such as reference answers or ground-truth regions of interest. This raises a fundamental question: \emph{where can informative asymmetry come from when nothing privileged is available?}
We answer this by inverting where the asymmetry comes from. Rather than adding privileged information to the teacher, we subtract information from the student.
This asymmetry creates the same effective learning signal \emph{for free} as a teacher with access to information unavailable to the student, without ground-truth annotations, rewards, or a separate stronger teacher model.
Building on this principle, we introduce \textbf{Self-Supervised Visual On-Policy Distillation} (\method), a simple yet effective method that constructs on-policy learning signals from asymmetric augmented views. \method{} distills the teacher's distribution conditioned on the original image on-policy into the student distribution conditioned on a strongly augmented view of the same image. 
We systematically explore a broad design space of visual augmentations and uncover that (1) \emph{asymmetry matters}: all four augmentation families improve performance, while symmetric self-distillation degrades it; (2) \emph{strength matters}: performance peaks at a moderate strength; and (3) \emph{the gap must remain task-consistent}: augmentations that completely remove the question-relevant evidence can induce large but uninformative discrepancies. 
Across six fine-grained perception benchmarks, \method{} improves Qwen3.5-4B from $70.7\%$ to $77.4\%$---above all open-source models compared, up to Qwen3-VL at $235$B, and surpasses GPT-5.4. While holding training data the same, it recovers $96\%$ of the improvement achieved by methods with privileged information. Website is at \url{https://williamium3000.github.io/s2vopd}.

\end{abstract}

\section{Introduction}
\label{sec:intro}
On-policy distillation (OPD) has been shown to be effective in post-training of both LLMs~\citep{agarwal2024onpolicy,zhao2026selfdistilled}  and VLMs~\citep{bousselham2025vold,liu2026visualadvantage}. 
%
It improves a student policy by providing dense teacher supervision along the student's own generated trajectories~\citep{agarwal2024onpolicy,zhao2026selfdistilled,sang2026opsdc}. 
Central to this paradigm is an informative \emph{teacher--student asymmetry}: the teacher must possess information or capability that enables it to provide a better predictive target than the student itself. 
Conventional OPD obtains this asymmetry from a larger or stronger teacher. 
On-policy self-distillation (OPSD) removes the need for a separate teacher model, but instead conditions the teacher on privileged information, such as reference solutions~\citep{zhao2026selfdistilled,penaloza2026privileged}, environmental feedback~\citep{hubotter2026reinforcement}, or ground-truth regions of interest~\citep{yuan2026visionopd,liu2026visualadvantage,yoon2026decomposed}.
In either case, the learning signal ultimately relies on a privileged advantage external to the model, which becomes increasingly difficult to obtain as model capabilities outpace the supervision humans can reliably provide. 


This raises a fundamental question:

\begin{center}
\emph{where can informative asymmetry come from when nothing privileged is available?}
\end{center}
We show that it does not necessarily come from giving the teacher more information at all. Instead, we invert the direction of the asymmetry: \emph{rather than adding information to the teacher, we withhold information from the student}. The teacher observes the original input, while the student operates on a degraded view of the same input. Their difference in available visual evidence induces a predictive discrepancy and thereby creates the asymmetry required for distillation---without ground-truth annotations, rewards, privileged context, or a separately trained teacher.
This idea is similar to self-supervised learning, where supervision is constructed from multiple views of the same unlabeled input~\citep{tian2020infomin,chen2021simsiam,tian2025beyond,zaher2026discrete,sharma2026self}.
Contrastive methods such as SimCLR~\citep{chen2020simclr} learn by aligning representations of differently augmented views, while teacher--student methods such as BYOL~\citep{grill2020byol} and DINO~\citep{caron2021dino} use one view to provide learning targets for another augmented view.
They all share the principle that augmentation defines invariances that allow models to learn without external labels. Our method follows the same principle, but uses augmentation for a different purpose. Rather than defining a representation-level objective across augmented views, we apply augmentation to invert where the
asymmetry comes from, thereby creating an informative asymmetry \emph{for free}. 

Building on this idea, we introduce \textbf{Self-Supervised Visual On-Policy Distillation} (\method{}). Given an image $x$ and question $q$, an exponentially moving average teacher observes the original image, while the student generates on-policy trajectories from an augmented view $\tilde{x}=T(x)$.
\method then conditions the teacher distribution on the original image and distills into the student trajectories generated from an augmented view $\tilde{x}=T(x)$. This transformation $T$ induces a predictive gap between teacher and student for free, without labels, regions of interest, or external rewards; the student-side transformation is the sole source of the teacher's informational advantage.

Once asymmetry is constructed this way, a more important question emerges: \emph{how to construct asymmetry through augmentation?}
We answer this question empirically, with what is to our knowledge the first controlled, large-scale search of the augmentation space for OPD. We show that: (1) \emph{asymmetry itself matters}: all four augmentation families improve while symmetric self-distillation degrades; (2) \emph{strength matters}: performance peaks at a moderate teacher--student gap; (3) \emph{the gap must remain task-consistent}: aggressive cropping enlarges the discrepancy but removes question-relevant evidence, making the supervision less useful.

These findings identify the most effective instantiation of \method{}: downscaling the student view to $0.3$--$0.6\times$ resolution with stochastic Gaussian noise. Across six fine-grained perception benchmarks, \method{} improves Qwen3.5-4B from $70.7\%$ to $77.4\%$ in average accuracy, a gain of $6.7\%$. This places a $4$B model above Qwen3-VL-Instruct-235B ($75.8\%$) and GPT-5.4 ($72.8\%$), and matches the much larger Qwen3.5-397B. Moreover, \method{} consistently improves both fine-grained perception and math reasoning, by $+5.7\%$ and $+3.7\%$ at $4$B ($+3.6\%$ and $+3.2\%$ at $9$B), surpassing both methods with privileged-supervision and self-rewarding RL methods.



\section{Related Work}
\label{sec:related}

\textbf{Self- and semi-supervised learning.}
A line of self-supervised representation learning derives supervision directly from augmented views of unlabeled images. Contrastive methods such as SimCLR~\citep{chen2020simclr} and MoCo~\citep{he2020moco} align different augmented views of the same image, while negative-free methods including BYOL~\citep{grill2020byol}, SimSiam~\citep{chen2021simsiam}, and DINO~\citep{caron2021dino} match a student to a stop-gradient or momentum teacher under different augmentations.  Similar asymmetry appears in semi-supervised consistency learning, where predictions from weakly augmented inputs supervise strongly augmented ones~\citep{laine2017temporal,tarvainen2017meanteacher,xie2020uda,sohn2020fixmatch,xie2020noisystudent, li2023diverse}. Across these paradigms, augmentation is not merely regularization but a central source of supervision, and its design strongly shapes what is learned~\citep{tian2020infomin,cubuk2019autoaugment,cubuk2020randaugment}. We adopt this principle for token-level generative distillation: the teacher observes the clean image, the student an augmented view, and their next-token distribution gap provides the supervision for free.

\textbf{Visual augmentation in vision-language models.}
Augmentation plays a much smaller role in vision-language model training than in representation learning, as geometric or photometric perturbations can alter instruction-relevant content such as text, object identity, and spatial relations.  A recent exception is reinforcement learning for visual reasoning: NoisyRollout~\cite{liu2025noisyrollout} uses perturbed images to diversify exploration, while VPPO~\cite{huang2025vppo} and PRPO~\cite{li2026prpo} compare policy distributions across clean and perturbed views to identify perception-critical tokens and reweight policy gradients. In all cases, augmentation modulates learning driven by external rewards, with both views evaluated by the same policy. In contrast, we apply augmentation only to the student: the discrepancy from a teacher observing the clean image is itself the training signal, requiring neither rewards nor annotations.

\textbf{Knowledge distillation and on-policy distillation.}
On-policy distillation (OPD)~\cite{agarwal2024onpolicy} trains the student on its own trajectories while receiving dense token-level supervision from a teacher, reducing the train--inference mismatch of offline distillation. Subsequent work studies improved divergence objectives, selective supervision over informative tokens and trajectories~\cite{liu2026visualadvantage}, and optimization strategies that better preserve visual grounding~\cite{yoon2026decomposed}. On-policy self-distillation (OPSD) extends this idea further and removes the need for a larger teacher by sharing a single model between student and teacher, with the teacher granted access to privileged information such as a ground-truth solution. Existing methods obtain this teacher-student asymmetry from verified reasoning traces~\cite{zhao2026selfdistilled}, additional context~\cite{ye2026context}, environmental feedback~\cite{hubotter2026reinforcement}, or privileged visual information~\cite{yuan2026visionopd,liu2026visualadvantage,yoon2026decomposed,sun2026vzero}. These signals, however, require external annotation or guidance that might be costly or unavailable. We mitigate this by inverting where the
asymmetry comes from. Rather than adding privileged information to the teacher, we subtract information from the student, by conditioning the student on a strongly augmented view of the same image.  Our focus is therefore the design of this augmentation transformation and the properties that make this teacher-student discrepancy an effective supervisory signal.

\section{Method}
\label{sec:method}

\subsection{Self-Supervised Visual On-policy Distillation}
\label{sec:method:opsd}
Given an image--question pair $(x,q)\sim\mathcal{D}$, let $\pi_\theta$ denote the policy to be optimized and $\pi_\phi$ its exponential-moving-average (EMA) as teacher:
$
\phi \leftarrow (1-\eta)\,\phi + \eta\,\theta$.
Unlike conventional on-policy distillation, $\pi_\phi$ is neither a separately trained stronger model nor conditioned on privileged supervision. Instead, we construct the teacher--student asymmetry directly from their visual inputs.

\textbf{Asymmetric on-policy views.}
For each $(x,q)\sim\mathcal{D}$, we sample a stochastic transformation $T \sim \mathcal{T}$ that produces the student view $\tilde{x} = T(x)$, while the teacher retains the clean view $x$. The student samples $n$ rollouts $y^{(1)}, \ldots, y^{(n)} \sim \pi_\theta(\cdot \mid \tilde{x}, q)$ conditioned on its corrupted view $\tilde{x}$. This gives us a teacher distribution conditioned on the clean view $x$: $p^{\tau}_t = \pi_\phi(\cdot \mid x, q, y_{<t})$ and the student distribution conditioned on the augmented view $\tilde{x}$: $p^{s}_t = \pi_\theta(\cdot \mid \tilde{x}, q, y_{<t})$.

\textbf{On-policy distillation.}
With original clean view $x$ for teacher model and the augmented view $\tilde{x}$ for the student, \method then conditions the
teacher's next-token distribution on the clean view $x$, and distills into the student's own trajectories generated on the degraded view $\tilde{x}$.
Let $D(\cdot\,\|\,\cdot)$ denote a divergence between two distributions over the vocabulary. The training objective minimizes the expected per-token divergence between the teacher and the student distributions, evaluated along trajectories sampled from the student:
\begin{equation}
\mathcal{L}(\theta) \;=\; \mathbb{E}_{(x,q)\sim\mathcal{D}}\;
\mathbb{E}_{y\sim\pi_\theta(\cdot\mid\tilde{x},q)}\!\left[\,\frac{1}{|y|}\sum_{t=1}^{|y|}
D\big(\pi_\phi\big(\cdot \mid x,\,q,\,y_{<t}\big) \,\big\|\, \pi_\theta\big(\cdot \mid \tilde{x},\,q,\,y_{<t}\big)\big)\right].
\label{eq:obj}
\end{equation}
The inner expectation renders the objective on-policy: the prefixes $y_{<t}$ at which the two distributions are compared are drawn from the student's own policy, so that supervision is provided precisely at the states the student visits at inference time.

We instantiate $D$ as the generalized Jensen--Shannon divergence $D^{\alpha}_{\mathrm{JS}}$, which interpolates between the forward and reverse Kullback--Leibler divergence $D_{\mathrm{KL}}$ through a mixture weight $\alpha$ and remains bounded when the two distributions have limited overlap. With $\alpha = 0.5$ and the mixture distribution $m_t = \alpha\, \pi_\phi(\cdot \mid x, q, y_{<t}) + (1-\alpha)\, \pi_\theta(\cdot \mid \tilde{x}, q, y_{<t})$,
\begin{equation}
\begin{aligned}
D^{\alpha}_{\mathrm{JS}}\big(\pi_\phi\big(\cdot \mid x,\,q,\,y_{<t}\big) \,\big\|\, \pi_\theta\big(\cdot \mid \tilde{x},\,q,\,y_{<t}\big)\big)
\;=\;& \alpha\, D_{\mathrm{KL}}\big(\pi_\phi\big(\cdot \mid x,\,q,\,y_{<t}\big) \,\big\|\, m_t\big) \\
+\;& (1-\alpha)\, D_{\mathrm{KL}}\big(\pi_\theta\big(\cdot \mid \tilde{x},\,q,\,y_{<t}\big) \,\big\|\, m_t\big).
\end{aligned}
\label{eq:jsd}
\end{equation}
Both distributions are restricted to the teacher's top-$k$ tokens at position $t$ and renormalized over this support, which stabilizes the objective against the long tail of the vocabulary.

Crucially, the transformation $T$ constructs an input asymmetry that induces a predictive discrepancy between the student distribution $p_t^{s}$ and the teacher distribution $p_t^{\tau}$. Through $T$, the student is trained on a deliberately degraded view of the same image, while the teacher evaluates the identical on-policy prefix under the original view.
Unlike conventional OPD, which obtains this asymmetry from a stronger teacher or privileged supervision, \method creates it by withholding visual information from the student. This yields an analogous on-policy distillation signal for free without external labels, rewards, privileged annotations, or a stronger teacher. The effectiveness of this self-constructed supervision therefore depends on how $T$ shapes the teacher--student predictive gap.

\begin{figure}
    \centering
    \includegraphics[width=1\linewidth]{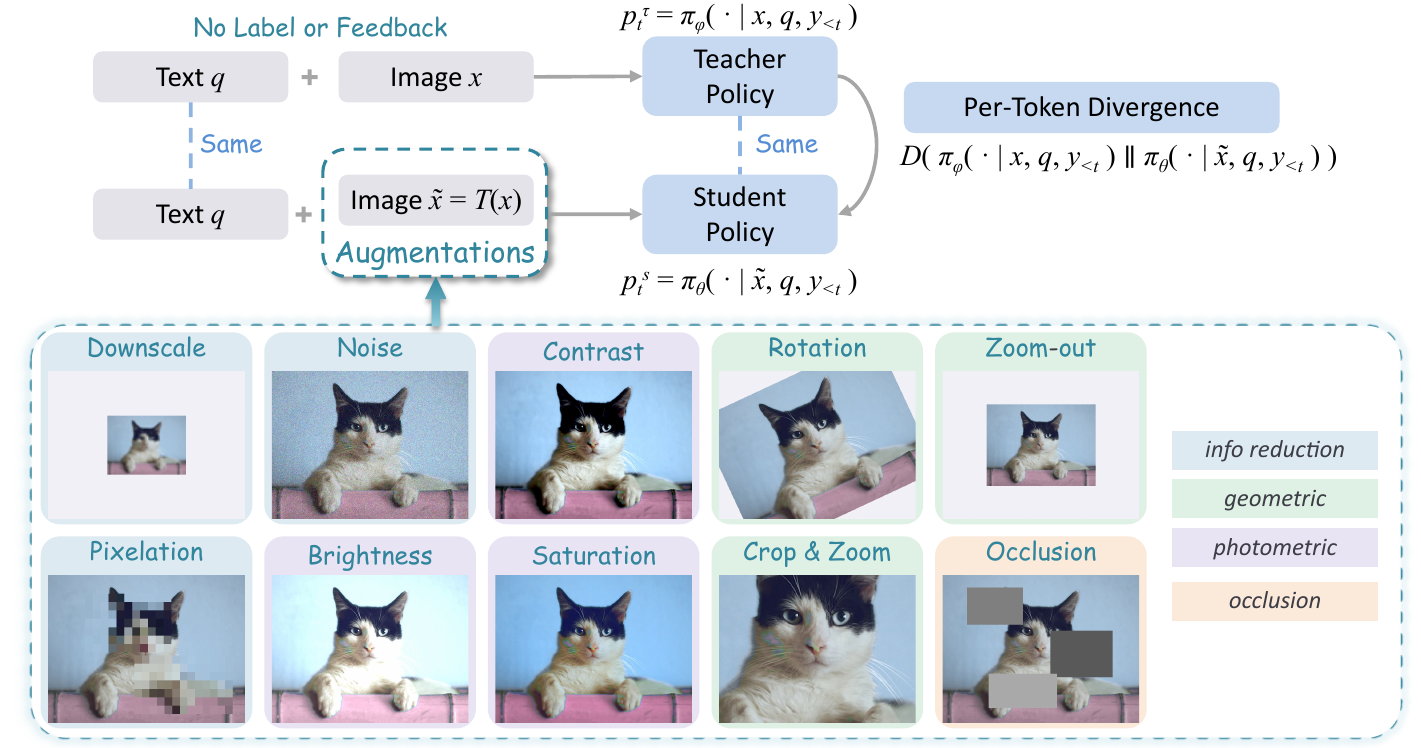}
    \caption{Overview of \method{}. The student generates rollouts from a corrupted view of the image; an EMA teacher scores the same generated prefixes under the clean view; a top-$k$ generalized JSD transfers the teacher's better-informed token distributions to the student. The augmentation applied to the student view is the sole source of supervision and the object of our study.}
    \label{fig:placeholder}
\end{figure}

\subsection{Constructing Asymmetry for Free}
\label{sec:method:aug}

\paragraph{Augmentation space.}
The transformation $T$ is specified by three components: a set of primitive operators together with their probabilities and strength ranges, a composition policy for combining selected operators, and a global probability controlling whether augmentation is applied at all. We organize these operators by how they alter the visual information available to the student, yielding four families (summarized in Table~\ref{tab:augspace}):
\begin{itemize}
\item \textbf{Information reduction.} Operators that reduce usable visual information while preserving spatial layout. Subtractive variants remove signal content, including downscaling, Gaussian blur, pixelation, and spectral band-stop filtering; additive variants reduce signal-to-noise ratio through Gaussian noise. We also include visual-token dropping, which reduces the number of encoder tokens while preserving the resolution and positions of those retained.

\item \textbf{Geometric.} Operators that modify spatial organization, including rotation, translation, cropping, and zoom-out with padding. Unlike the other families, these transformations may alter the answer to spatially grounded questions by changing object locations relative to the image frame.

\item \textbf{Photometric.} Operators that modify appearance while preserving geometry and content, including brightness, contrast, saturation, hue, gamma, sharpness, and histogram equalization. These transformations leave object boundaries and spatial relations unchanged.

\item \textbf{Occlusion.} Operators that remove localized image regions while leaving the remainder unchanged, including random erasing, grid masking, and filled crops. This family is distinguished from information reduction by spatial locality. We additionally vary the fill content (e.g., constant or noise) and occlusion granularity (few large versus many small regions).
\end{itemize}

\begin{table}[t]
\caption{Overview of augmentation operators on the student policy.}
\label{tab:augspace}
\centering
\footnotesize
\setlength{\tabcolsep}{4.5pt}
\begin{tabular}{@{}lll@{}}
\toprule
\textbf{Operator} & \textbf{Effects} & \textbf{Strengths Range} \\
\midrule
\multicolumn{3}{@{}l}{\textcolor{hdrblue}{\emph{Information reduction}}} \\
Downscale & fewer pixels and visual tokens (no resize-back) & $s\,{\sim}\,\mathcal{U}(0.3,0.6)$; $0.4$--$0.7$; $0.25$--$0.5$ \\
Downscale, 3-tier & one of three bands drawn per sample & $[0.2,0.35]$, $[0.35,0.5]$, $[0.5,0.75]$ \\
Gaussian blur & soft low-pass & radius $\mathcal{U}(0.5,1.5)$; $(1.5,3.0)$; $(3.0,6.0)$ \\
Spectral band-stop & removes an annulus of the 2D spectrum & inner radius $0.15$--$0.35$, width $0.15$--$0.3$ \\
Gaussian noise & additive; lowers signal-to-noise ratio & DDPM step $200$ ($\sigma{\approx}0.11$), $\rho{=}0.5$ \\
Local pixelation & piecewise-constant low-pass (mosaic patches) &  application probability $\rho{=}0.4$ \\
Visual-token drop & fewer encoder tokens at native fidelity & drop $15/30/50\%$ \\
\midrule
\multicolumn{3}{@{}l}{\textcolor{hdrblue}{\emph{Geometric}}} \\
Rotation & rotates the frame & $\pm 35^\circ$ \\
Translation & shifts the frame & up to $15\%$ of the side length \\
Crop & removes peripheral context, and sometimes the target & min.\ scale $0.7$ / $0.5$ / $0.3$ \\
Zoom-out & shrinks the scene onto a padded canvas & scale $\mathcal{U}(0.4,0.8)$ \\
\midrule
\multicolumn{3}{@{}l}{\textcolor{hdrblue}{\emph{Photometric}}} \\
Colour jitter & brightness / contrast / saturation, $\rho{=}0.8$ each & $0.5$--$1.5$\,/\,$0.5$--$1.8$\,/\,$0.2$--$1.8$ \\
Heavy photometric & increase hue, gamma, sharpness, $\rho{=}0.5$ each & hue $\pm0.15$; gamma $0.6$--$1.6$; sharp.\ $0.2$--$2.5$ \\
\midrule
\multicolumn{3}{@{}l}{\textcolor{hdrblue}{\emph{Occlusion}}} \\
Random erasing, grey fill & deletes patches, constant fill & ${\leq}3$ patches, $5$--$30\%$ area \\
Random erasing, noise fill & deletes patches, random fill & ${\leq}3$ patches, $5$--$30\%$ area \\
GridMask & regular-lattice masking, grey fill & cell period $0.1$--$0.3$, keep ratio $0.5$--$0.7$ \\
\bottomrule
\end{tabular}
\end{table}

Formally, let $\mathcal{A}=\{A_1,\ldots,A_M\}$ denote the collection of candidate operators, where operator $A_j$ is associated with an application probability $\rho_j$ and a strength $\lambda_j\sim P_j$. We first sample a global augmentation indicator
\begin{equation}
z \sim \operatorname{Bernoulli}(p),
\label{eq:coverage}
\end{equation}
where $p$ controls the fraction of student views that are augmented. When $z=0$, the student receives the original view, i.e., $T(x)=x$. When $z=1$, we independently sample
\begin{equation}
b_j \sim \operatorname{Bernoulli}(\rho_j),
\qquad j=1,\ldots,M,
\label{eq:opselect}
\end{equation}
and compose the selected operators in a fixed order:
\begin{equation}
T(x)
=
\begin{cases}
x, & z=0, \\[2mm]
\left(
A_M^{b_M}(\,\cdot\,;\lambda_M)
\circ \cdots \circ
A_1^{b_1}(\,\cdot\,;\lambda_1)
\right)(x),
& z=1,
\end{cases}
\label{eq:aug}
\end{equation}
where $A_j^{0}$ denotes the identity and $A_j^{1}$ applies $A_j$. This formulation jointly controls the augmentation family, the strength, the per-operator probability, the composition, and the probability $1-p$ of retaining an unaugmented view, and thereby allows us to construct teacher--student asymmetries of different forms and magnitudes in a systematic way. Throughout, augmentation is applied to the student view only; the teacher always observes the original image.

\paragraph{Best recipe.}
 Among the configurations we explored, the strongest composes two operators of the information-reduction family: downscaling followed by additive Gaussian noise. We set $p=1$, so that every student view is augmented, and define
\begin{equation}
T(x)
=
A_{\mathrm{noise}}
\left(
A_{\mathrm{down}}(x;s);
\lambda_{\mathrm{noise}}
\right),
\qquad
s\sim\mathcal{U}(0.3,0.6),
\label{eq:best}
\end{equation}
where $A_{\mathrm{down}}$ downsamples the image without resizing it back to the original resolution, leading to fewer visual tokens input to the student. Downscaling is applied to every sample, whereas Gaussian noise is applied independently with probability $\rho_{\mathrm{noise}}=0.5$; the noise follows a DDPM forward step, $x \mapsto \sqrt{\bar\alpha_t}\,x + \sqrt{1-\bar\alpha_t}\,\varepsilon$ at $t{=}200$, equivalent to additive Gaussian noise of standard deviation $\approx 0.11$ with negligible signal attenuation. As a practical benefit, lower-resolution student inputs reduce both rollout and student forward-pass costs. We use this composition as the default instantiation of $T$ in \method{}.


\section{Experiments}
\label{sec:exp}

\subsection{Setup}
\label{sec:exp:setup}

\textbf{Model and training.}
We adopt Qwen3.5-4B and Qwen3.5-9B~\citep{qwen3.5} as base models for all experiments. We utilize vLLM~\citep{kwon2023vllm} as the inference engine for rollouts. Every run uses an identical configuration, unless specified otherwise: batch size $96$ prompts, $n{=}8$ rollouts per prompt, learning rate $2\times 10^{-6}$ with $10$ warmup steps, $65$ / $130$ optimizer steps (one epoch of the $6$K / $12$K training samples). We apply EMA with $\eta{=}0.05$ on the teacher and a maximum prompt length of $8{,}192$ and response length of $1{,}024$ tokens.

We sample 12K questions from the natural-image domain of FineVision~\citep{wiedmann2025finevision} as our training data and train for 130 optimizer steps. For a fair comparison, we also adopt Vision-OPD-6K~\citep{yuan2026visionopd}, whose ground-truth region annotations are only provided to Vision-OPD~\citep{yuan2026visionopd} for training and are never used by \method{}.

\textbf{Evaluation.}
We evaluate on six perception benchmarks and three math reasoning benchmarks: V*Bench~\citep{wu2024vstar}, ZoomBench~\citep{wei2026zooming}, HR-Bench~4K and 8K~\citep{wang2024hrbench}, MME-RealWorld and its Chinese subset~\citep{zhang2024mmerealworld}, MathVista~\citep{lu2024mathvista}, MathVerse~\citep{zhang2024mathverse}, and MathVision~\citep{wang2024mathvision}. 
For inference on the six perception benchmarks, we use greedy decoding with a maximum of $4{,}096$ tokens; for math reasoning benchmarks, we adopt a $24{,}576$-token budget with temperature $T{=}0.3$, top-$p$ $0.95$, top-$k$ $20$, and presence penalty $1.5$. Generations are evaluated by first extracting the final answer and applying near-exact matching; only cases unresolved by these rules are adjudicated by an LLM judge (i.e., Qwen2.5-72B-Instruct).

\textbf{Baselines.} We compare \method with various models and baselines:
(i) the base model; (ii) symmetric self-distillation (w/o Aug. in Table~\ref{tab:component}) where on-policy self-distillation is performed without privileged information and augmentation; (iii) three methods with different privileged information, i.e.,  GT region of interest annotations (ZwZ~\citep{wei2026zooming}, Vision-OPD~\citep{yuan2026visionopd}), and  \opsd{} with GT answer as privileged information for teacher. To apply \opsd{} to vision domain, we train \opsd{} with Qwen3.5-4B on the Vision-OPD-6k dataset; and (iv) three self-rewarding RL methods, i.e., TTRL~\citep{zuo2025ttrl}, Intuitor~\citep{zhao2025intuitor}, and RENT~\citep{prabhudesai2025rent}, all trained for 65 steps on Vision-OPD-6k dataset, following a similar configurations. 

\subsection{Main results}
\label{sec:exp:main}

\begin{table}[t]
\caption{Performance of \method{} trained on a 12k subset of FineVision~\citep{wiedmann2025finevision}, compared with open-source, proprietary, and methods that use privileged information on fine-grained perception benchmarks (accuracy, \%; \avg{} is the average over the 6 benchmarks).}
\label{tab:fv}
\centering
\small
\setlength{\tabcolsep}{5pt}
\resizebox{0.99\linewidth}{!}{%
\begin{tabular}{lc|cccccc|c}
\toprule
\textbf{Method} & \textbf{Size} & \textbf{V*} & \textbf{Zoom} & \textbf{HR-4K} & \textbf{HR-8K} & \textbf{MME-RW} & \textbf{MME-RW-CN} & \textbf{\avg{}} \\
\hline
\multicolumn{2}{l|}{\rule{0pt}{2.5ex}\textcolor{hdrblue}{\emph{Open-source models}}} & & & & & & & \\
MiniCPM-V-4.5 & -- & \cellcolor{cellblue!10}70.68 & \cellcolor{cellblue!16}42.60 & \cellcolor{cellblue!16}69.63 & \cellcolor{cellblue!8}61.50 & \cellcolor{cellblue!16}62.65 & \cellcolor{cellblue!32}61.64 & \cellcolor{cellblue!8}61.45 \\
Qwen2.5-VL & 7B & \cellcolor{cellblue!35}78.53 & \cellcolor{cellblue!16}42.49 & \cellcolor{cellblue!23}71.62 & \cellcolor{cellblue!26}67.88 & \cellcolor{cellblue!8}60.80 & \cellcolor{cellblue!19}58.30 & \cellcolor{cellblue!15}63.27 \\
MiMo-VL-RL & 7B & \cellcolor{cellblue!50}83.25 & \cellcolor{cellblue!26}45.68 & \cellcolor{cellblue!29}73.50 & \cellcolor{cellblue!30}69.38 & \cellcolor{cellblue!17}62.73 & \cellcolor{cellblue!9}55.89 & \cellcolor{cellblue!23}65.07 \\
Qwen3-VL & 4B & \cellcolor{cellblue!40}80.10 & \cellcolor{cellblue!8}40.24 & \cellcolor{cellblue!44}78.25 & \cellcolor{cellblue!40}72.88 & \cellcolor{cellblue!20}63.47 & \cellcolor{cellblue!40}63.63 & \cellcolor{cellblue!28}66.43 \\
Thyme & 7B & \cellcolor{cellblue!47}82.20 & \cellcolor{cellblue!24}45.09 & \cellcolor{cellblue!40}77.00 & \cellcolor{cellblue!38}72.00 & \cellcolor{cellblue!26}64.80 & \cellcolor{cellblue!43}64.59 & \cellcolor{cellblue!33}67.61 \\
DeepEyesV2 & 7B & \cellcolor{cellblue!45}81.68 & \cellcolor{cellblue!24}44.97 & \cellcolor{cellblue!43}77.88 & \cellcolor{cellblue!43}73.75 & \cellcolor{cellblue!27}64.90 & \cellcolor{cellblue!45}65.07 & \cellcolor{cellblue!35}68.04 \\
DeepEyes & 7B & \cellcolor{cellblue!58}85.86 & \cellcolor{cellblue!29}46.51 & \cellcolor{cellblue!34}75.13 & \cellcolor{cellblue!40}72.63 & \cellcolor{cellblue!23}64.10 & \cellcolor{cellblue!41}64.09 & \cellcolor{cellblue!35}68.05 \\
Qwen3-VL-Instruct & 8B & \cellcolor{cellblue!55}84.82 & \cellcolor{cellblue!17}42.96 & \cellcolor{cellblue!48}79.63 & \cellcolor{cellblue!47}75.25 & \cellcolor{cellblue!19}63.19 & \cellcolor{cellblue!43}64.61 & \cellcolor{cellblue!36}68.41 \\
GLM-4.5V & -- & \cellcolor{cellblue!50}83.25 & \cellcolor{cellblue!39}49.23 & \cellcolor{cellblue!55}81.63 & \cellcolor{cellblue!46}74.88 & \cellcolor{cellblue!32}66.04 & \cellcolor{cellblue!28}60.71 & \cellcolor{cellblue!40}69.29 \\
Qwen3.5 \textcolor{hdrblue}{(our base)} & 4B & \cellcolor{cellblue!53}84.29 & \cellcolor{cellblue!33}47.69 & \cellcolor{cellblue!63}84.38 & \cellcolor{cellblue!61}80.13 & \cellcolor{cellblue!22}63.86 & \cellcolor{cellblue!40}63.70 & \cellcolor{cellblue!45}70.68 \\
GLM-4.6V & 106B & \cellcolor{cellblue!62}86.91 & \cellcolor{cellblue!41}50.06 & \cellcolor{cellblue!56}82.13 & \cellcolor{cellblue!57}78.88 & \cellcolor{cellblue!30}65.57 & \cellcolor{cellblue!47}65.62 & \cellcolor{cellblue!49}71.53 \\
Kimi-K2.6 & 1T & \cellcolor{cellblue!67}88.48 & \cellcolor{cellblue!52}53.14 & \cellcolor{cellblue!55}81.88 & \cellcolor{cellblue!55}78.00 & \cellcolor{cellblue!47}69.22 & \cellcolor{cellblue!49}66.13 & \cellcolor{cellblue!54}72.81 \\
SenseNova-MARS & 8B & \cellcolor{cellblue!78}92.15 & \cellcolor{cellblue!34}47.81 & \cellcolor{cellblue!59}83.13 & \cellcolor{cellblue!56}78.38 & \cellcolor{cellblue!40}67.90 & \cellcolor{cellblue!60}68.90 & \cellcolor{cellblue!55}73.05 \\
Kimi-K2.5 & 1T & \cellcolor{cellblue!58}85.86 & \cellcolor{cellblue!63}56.33 & \cellcolor{cellblue!55}81.87 & \cellcolor{cellblue!47}75.38 & \cellcolor{cellblue!57}71.51 & \cellcolor{cellblue!58}68.40 & \cellcolor{cellblue!56}73.23 \\
Qwen3-VL-Instruct & 235B & \cellcolor{cellblue!75}91.10 & \cellcolor{cellblue!62}56.09 & \cellcolor{cellblue!69}86.13 & \cellcolor{cellblue!62}80.38 & \cellcolor{cellblue!58}71.74 & \cellcolor{cellblue!61}69.04 & \cellcolor{cellblue!66}75.75 \\
Qwen3.5 & 397B & \cellcolor{cellblue!65}87.96 & \cellcolor{cellblue!66}57.16 & \cellcolor{cellblue!79}89.38 & \cellcolor{cellblue!76}85.50 & \cellcolor{cellblue!72}74.82 & \cellcolor{cellblue!64}69.82 & \cellcolor{cellblue!73}77.44 \\
\multicolumn{2}{l|}{\rule{0pt}{2.5ex}\textcolor{hdrblue}{\emph{Proprietary models}}} & & & & & & & \\
GPT-5.1 & -- & \cellcolor{cellblue!8}70.16 & \cellcolor{cellblue!32}47.22 & \cellcolor{cellblue!8}67.00 & \cellcolor{cellblue!19}65.25 & \cellcolor{cellblue!23}64.04 & \cellcolor{cellblue!8}55.57 & \cellcolor{cellblue!8}61.54 \\
GPT-5.2 & -- & \cellcolor{cellblue!36}79.06 & \cellcolor{cellblue!44}50.89 & \cellcolor{cellblue!53}81.12 & \cellcolor{cellblue!56}78.38 & \cellcolor{cellblue!62}72.60 & \cellcolor{cellblue!60}68.80 & \cellcolor{cellblue!50}71.81 \\
GPT-5.4 & -- & \cellcolor{cellblue!30}76.96 & \cellcolor{cellblue!50}52.66 & \cellcolor{cellblue!62}84.00 & \cellcolor{cellblue!54}77.88 & \cellcolor{cellblue!69}74.20 & \cellcolor{cellblue!68}70.93 & \cellcolor{cellblue!54}72.77 \\
Gemini-3-Flash & -- & \cellcolor{cellblue!60}86.39 & \cellcolor{cellblue!73}59.29 & \cellcolor{cellblue!74}87.88 & \cellcolor{cellblue!75}85.00 & \cellcolor{cellblue!72}74.86 & \cellcolor{cellblue!75}72.62 & \cellcolor{cellblue!74}77.67 \\
Gemini-3.5-Flash & -- & \cellcolor{cellblue!68}89.01 & \cellcolor{cellblue!80}61.42 & \cellcolor{cellblue!78}89.12 & \cellcolor{cellblue!79}86.62 & \cellcolor{cellblue!74}75.31 & \cellcolor{cellblue!80}73.97 & \cellcolor{cellblue!80}79.24 \\
Gemini-3.1-Pro & -- & \cellcolor{cellblue!65}87.96 & \cellcolor{cellblue!79}61.18 & \cellcolor{cellblue!80}89.63 & \cellcolor{cellblue!80}86.88 & \cellcolor{cellblue!80}76.53 & \cellcolor{cellblue!77}73.31 & \cellcolor{cellblue!80}79.25 \\
\hline
\multicolumn{2}{l|}{\rule{0pt}{2.5ex}\textcolor{hdrblue}{\emph{w/ privileged info.}}} & & & & & & & \\
ZwZ \textcolor{hdrblue}{(Qwen2.5-VL-7B)} & 7B & \cellcolor{cellblue!67}88.48 & \cellcolor{cellblue!60}55.62 & \cellcolor{cellblue!35}75.38 & \cellcolor{cellblue!41}73.25 & \cellcolor{cellblue!33}66.21 & \cellcolor{cellblue!53}66.96 & \cellcolor{cellblue!47}70.98 \\
ZwZ \textcolor{hdrblue}{(Qwen3-VL-4B)} & 4B & \cellcolor{cellblue!80}92.67 & \cellcolor{cellblue!61}55.74 & \cellcolor{cellblue!55}81.75 & \cellcolor{cellblue!59}79.50 & \cellcolor{cellblue!43}68.52 & \cellcolor{cellblue!57}68.09 & \cellcolor{cellblue!60}74.38 \\
ZwZ \textcolor{hdrblue}{(Qwen3-VL-8B)} & 8B & \cellcolor{cellblue!75}91.10 & \cellcolor{cellblue!69}58.11 & \cellcolor{cellblue!63}84.38 & \cellcolor{cellblue!66}82.00 & \cellcolor{cellblue!50}69.87 & \cellcolor{cellblue!67}70.59 & \cellcolor{cellblue!67}76.01 \\
\opsd{} \textcolor{hdrblue}{(trained on Vision-OPD-6K)} & 4B & \cellcolor{cellblue!45}81.68 & \cellcolor{cellblue!50}52.54 & \cellcolor{cellblue!53}81.25 & \cellcolor{cellblue!54}77.75 & \cellcolor{cellblue!59}71.91 & \cellcolor{cellblue!72}71.96 & \cellcolor{cellblue!54}72.85 \\
Vision-OPD & 4B & \cellcolor{cellblue!78}92.15 & \cellcolor{cellblue!74}59.76 & \cellcolor{cellblue!64}84.50 & \cellcolor{cellblue!62}80.38 & \cellcolor{cellblue!72}74.88 & \cellcolor{cellblue!67}70.76 & \cellcolor{cellblue!71}77.07 \\
\multicolumn{2}{l|}{\rule{0pt}{2.5ex}\textcolor{hdrblue}{\emph{w/o privileged info.\ }}} & & & & & & & \\
\method{} (ours) & 4B & \cellcolor{cellblue!76}91.48 & \cellcolor{cellblue!62}55.98 & \cellcolor{cellblue!70}86.38 & \cellcolor{cellblue!66}82.00 & \cellcolor{cellblue!78}76.13 & \cellcolor{cellblue!75}72.66 & \cellcolor{cellblue!73}77.44 \\
\bottomrule
\end{tabular}%
}
\end{table}

Table~\ref{tab:fv} reports the performance of \method{} trained on a 12K subset of FineVision~\citep{wiedmann2025finevision}, compared against open-source models, proprietary models, and methods trained with privileged information on six perception benchmarks. \method{} improves Qwen3.5-4B from $70.68\%$ to $77.44\%$ average accuracy, a gain of $6.76\%$. This delivers a $4$B model that outperforms all open-source models in Table~\ref{tab:fv}, including Qwen3-VL-Instruct-235B ($75.75\%$), and matches the performance of Qwen3.5-397B ($77.44\%$). Comparing with proprietary models, \method{} outperforms the GPT-5 series and is on par with Gemini-3-Flash, trailing only Gemini-3.5-Flash and Gemini-3.1-Pro by less than $2\%$. It also surpasses all privileged-supervised methos reported, including ZwZ, Vision-OPD, and \opsd{}, despite using no ground-truth regions or answers during training.

\paragraph{Fair comparison with prior methods.}
To isolate the effect of training data from that of our method, we further evaluate \method{} trained on Vision-OPD-6K~\citep{yuan2026visionopd} and compare it with prior privileged-information on-policy distillation methods and self-rewarding RL methods. All models in Table~\ref{tab:main} are trained on Vision-OPD-6K for 65 steps using the same training configuration and evaluation protocols. For all methods except the three self-rewarding RL baselines, we report performance from the final checkpoint. The self-rewarding RL methods exhibit substantial training instability and eventually collapse, with some degrading to near-chance performance. We therefore report their best performance across 13 checkpoints saved at 5-step intervals, making the comparison conservative in their favor.

Table~\ref{tab:main} shows that \method{} achieves the best overall average among all methods without privileged information at both model scales: it improves over the strongest self-rewarding baseline by $+2.0\%$ at $4$B and by $+0.7\%$ at $9$B. \method{} also exceeds all supervised baselines, including ZwZ, Vision-OPD, and \opsd{} at the $4$B scale; at $9$B, it matches Vision-OPD as the runner-up overall while requiring no privileged annotations.

Interestingly, methods with privileged visual information substantially improve perception performance, consistent with their training on perception-oriented data, but are less effective for math reasoning. For example, \opsd{} decreases MathVision performance by $9.3\%$ at 4B and $7.9\%$ at 9B, while ZwZ reduces MathVerse accuracy by $27.1\%$ at 4B relative to the base model. In contrast, self-rewarding RL methods such as TTRL, Intuitor, and RENT generally improve math reasoning, potentially due to the longer and more deliberative responses induced by GRPO. However, their gains on perception remain  limited, likely because their optimization does not explicitly steer the model toward perception-oriented behavior, as Vision-OPD does.
\method{} achieves the best of both. It derives a perception-aligned learning signal from the model itself, without privileged annotations, while improving both visual perception and mathematical reasoning, achieving near-best perception performance while matching the strongest results on math reasoning.

\begin{table}[t]
\caption{Comparison with prior methods trained on the Vision-OPD training data, across six fine-grained perception and three mathematical-reasoning benchmarks at different model scales. \best{Bold} and \second{underline} mark the best and the second-best entry in each column within each base model; $^{\S}$ indicates that best instead of last checkpoint performance is reported.}
\label{tab:main}
\centering
\small
\setlength{\tabcolsep}{3pt}
\setlength{\belowrulesep}{0pt}
\setlength{\aboverulesep}{0pt}
\resizebox{\linewidth}{!}{%
\begin{tabular}{l*{6}{c}|*{3}{c}|c}
\toprule
& \multicolumn{6}{c|}{\rule{0pt}{2.4ex}\textbf{Perception}} & \multicolumn{3}{c|}{\textbf{Math reasoning}} & \\
\cline{2-7}\cline{8-10}
\rule{0pt}{2.4ex}\textbf{Method} & \textbf{V*} & \textbf{Zoom} & \textbf{HR-4K} & \textbf{HR-8K} & \textbf{MME-RW} & \textbf{MME-RW-CN} & \textbf{MathVista} & \textbf{MathVerse} & \textbf{MathVision} & \textbf{\avg{}} \\
\hline
\multicolumn{11}{c}{\rule{0pt}{2.6ex}Qwen3.5-4B\rule[-1.1ex]{0pt}{0pt}} \\
\hline
Qwen3.5-4B & \cellcolor{cellblue!30}84.29 & \cellcolor{cellblue!8}47.69 & \cellcolor{cellblue!68}\second{84.38} & \cellcolor{cellblue!43}80.13 & \cellcolor{cellblue!35}63.86 & \cellcolor{cellblue!30}63.70 & \cellcolor{cellblue!37}75.80 & \cellcolor{cellblue!68}67.56 & \cellcolor{cellblue!73}65.26 & \cellcolor{cellblue!32}70.30 \\
\multicolumn{7}{l|}{\rule{0pt}{2.5ex}\textcolor{hdrblue}{\emph{w/ privileged info.}}} & \multicolumn{3}{c|}{} & \\
ZwZ-4B\citep{wei2026zooming} & \cellcolor{cellblue!80}\best{92.67} & \cellcolor{cellblue!50}55.74 & \cellcolor{cellblue!8}81.75 & \cellcolor{cellblue!37}79.50 & \cellcolor{cellblue!54}68.52 & \cellcolor{cellblue!52}68.09 & \cellcolor{cellblue!8}71.80 & \cellcolor{cellblue!8}40.48 & \cellcolor{cellblue!8}52.17 & \cellcolor{cellblue!8}67.86 \\
\opsd{}~\citep{zhao2026selfdistilled} & \cellcolor{cellblue!39}85.86 & \cellcolor{cellblue!72}\second{59.88} & \cellcolor{cellblue!66}84.25 & \cellcolor{cellblue!8}76.75 & \cellcolor{cellblue!80}\best{74.70} & \cellcolor{cellblue!80}\best{73.40} & \cellcolor{cellblue!15}72.80 & \cellcolor{cellblue!64}66.17 & \cellcolor{cellblue!27}55.92 & \cellcolor{cellblue!50}72.19 \\
Vision-OPD~\citep{yuan2026visionopd} & \cellcolor{cellblue!71}\second{91.10} & \cellcolor{cellblue!80}\best{61.42} & \cellcolor{cellblue!17}82.12 & \cellcolor{cellblue!46}80.38 & \cellcolor{cellblue!79}\second{74.38} & \cellcolor{cellblue!62}69.98 & \cellcolor{cellblue!63}79.40 & \cellcolor{cellblue!77}71.68 & \cellcolor{cellblue!59}62.50 & \cellcolor{cellblue!75}\second{74.77} \\
\hdashline
\multicolumn{7}{l|}{\rule{0pt}{2.5ex}\textcolor{hdrblue}{\emph{w/o privileged info.}}} & \multicolumn{3}{c|}{} & \\
Intuitor$^{\S}$~\citep{zhao2025intuitor} & \cellcolor{cellblue!11}81.15 & \cellcolor{cellblue!27}51.24 & \cellcolor{cellblue!54}83.75 & \cellcolor{cellblue!67}\second{82.38} & \cellcolor{cellblue!8}57.50 & \cellcolor{cellblue!8}59.57 & \cellcolor{cellblue!80}\best{81.80} & \cellcolor{cellblue!77}71.98 & \cellcolor{cellblue!73}65.23 & \cellcolor{cellblue!34}70.51 \\
RENT$^{\S}$~\citep{prabhudesai2025rent} & \cellcolor{cellblue!8}80.63 & \cellcolor{cellblue!25}51.01 & \cellcolor{cellblue!34}82.88 & \cellcolor{cellblue!59}81.62 & \cellcolor{cellblue!12}58.55 & \cellcolor{cellblue!15}60.93 & \cellcolor{cellblue!70}80.47 & \cellcolor{cellblue!79}\second{72.64} & \cellcolor{cellblue!80}\best{66.74} & \cellcolor{cellblue!35}70.61 \\
TTRL$^{\S}$~\citep{zuo2025ttrl} & \cellcolor{cellblue!40}85.96 & \cellcolor{cellblue!50}55.62 & \cellcolor{cellblue!37}83.00 & \cellcolor{cellblue!62}81.88 & \cellcolor{cellblue!60}69.85 & \cellcolor{cellblue!39}65.56 & \cellcolor{cellblue!73}80.80 & \cellcolor{cellblue!77}71.68 & \cellcolor{cellblue!73}\second{65.33} & \cellcolor{cellblue!60}73.30 \\
\method{} (ours) & \cellcolor{cellblue!49}87.43 & \cellcolor{cellblue!62}57.99 & \cellcolor{cellblue!80}\best{84.88} & \cellcolor{cellblue!80}\best{83.62} & \cellcolor{cellblue!72}72.87 & \cellcolor{cellblue!69}\second{71.29} & \cellcolor{cellblue!78}\second{81.50} & \cellcolor{cellblue!80}\best{73.22} & \cellcolor{cellblue!72}65.13 & \cellcolor{cellblue!80}\best{75.33} \\
\hline
\multicolumn{11}{c}{\rule{0pt}{2.6ex}Qwen3.5-9B\rule[-1.1ex]{0pt}{0pt}} \\
\hline
Qwen3.5-9B & \cellcolor{cellblue!8}82.72 & \cellcolor{cellblue!8}52.07 & \cellcolor{cellblue!74}\second{85.75} & \cellcolor{cellblue!8}80.63 & \cellcolor{cellblue!33}71.40 & \cellcolor{cellblue!8}67.67 & \cellcolor{cellblue!42}78.80 & \cellcolor{cellblue!60}70.25 & \cellcolor{cellblue!65}66.91 & \cellcolor{cellblue!26}72.91 \\
\multicolumn{7}{l|}{\rule{0pt}{2.5ex}\textcolor{hdrblue}{\emph{w/ privileged info.}}} & \multicolumn{3}{c|}{} & \\
ZwZ-8B~\citep{wei2026zooming} & \cellcolor{cellblue!80}\best{91.10} & \cellcolor{cellblue!46}58.11 & \cellcolor{cellblue!40}84.38 & \cellcolor{cellblue!30}82.00 & \cellcolor{cellblue!8}69.87 & \cellcolor{cellblue!54}70.59 & \cellcolor{cellblue!16}76.00 & \cellcolor{cellblue!8}56.45 & \cellcolor{cellblue!8}57.01 & \cellcolor{cellblue!8}71.72 \\
\opsd{}~\citep{zhao2026selfdistilled} & \cellcolor{cellblue!48}87.43 & \cellcolor{cellblue!66}\second{61.18} & \cellcolor{cellblue!49}84.75 & \cellcolor{cellblue!28}81.88 & \cellcolor{cellblue!77}\second{74.10} & \cellcolor{cellblue!77}\second{72.00} & \cellcolor{cellblue!8}75.10 & \cellcolor{cellblue!59}70.10 & \cellcolor{cellblue!19}58.98 & \cellcolor{cellblue!41}73.95 \\
Vision-OPD~\citep{yuan2026visionopd} & \cellcolor{cellblue!62}89.01 & \cellcolor{cellblue!80}\best{63.43} & \cellcolor{cellblue!80}\best{86.00} & \cellcolor{cellblue!80}\best{85.12} & \cellcolor{cellblue!9}69.95 & \cellcolor{cellblue!28}68.92 & \cellcolor{cellblue!80}\best{83.00} & \cellcolor{cellblue!76}74.64 & \cellcolor{cellblue!77}68.98 & \cellcolor{cellblue!80}\best{76.56} \\
\hdashline
\multicolumn{7}{l|}{\rule{0pt}{2.5ex}\textcolor{hdrblue}{\emph{w/o privileged info.}}} & \multicolumn{3}{c|}{} & \\
RENT$^{\S}$~\citep{prabhudesai2025rent} & \cellcolor{cellblue!57}88.48 & \cellcolor{cellblue!37}56.69 & \cellcolor{cellblue!8}83.12 & \cellcolor{cellblue!28}81.88 & \cellcolor{cellblue!46}72.20 & \cellcolor{cellblue!11}67.89 & \cellcolor{cellblue!44}79.00 & \cellcolor{cellblue!77}74.92 & \cellcolor{cellblue!79}\second{69.31} & \cellcolor{cellblue!54}74.83 \\
Intuitor$^{\S}$~\citep{zhao2025intuitor} & \cellcolor{cellblue!62}89.01 & \cellcolor{cellblue!38}56.88 & \cellcolor{cellblue!74}\second{85.75} & \cellcolor{cellblue!54}\second{83.50} & \cellcolor{cellblue!45}72.18 & \cellcolor{cellblue!12}67.91 & \cellcolor{cellblue!57}80.50 & \cellcolor{cellblue!80}\second{75.56} & \cellcolor{cellblue!80}\best{69.57} & \cellcolor{cellblue!66}75.65 \\
TTRL$^{\S}$~\citep{zuo2025ttrl} & \cellcolor{cellblue!53}87.96 & \cellcolor{cellblue!34}56.14 & \cellcolor{cellblue!46}84.62 & \cellcolor{cellblue!22}81.50 & \cellcolor{cellblue!57}72.90 & \cellcolor{cellblue!14}68.08 & \cellcolor{cellblue!47}79.40 & \cellcolor{cellblue!77}74.72 & \cellcolor{cellblue!69}67.62 & \cellcolor{cellblue!53}74.77 \\
\method{} (ours) & \cellcolor{cellblue!76}\second{90.58} & \cellcolor{cellblue!39}56.92 & \cellcolor{cellblue!58}85.12 & \cellcolor{cellblue!38}82.50 & \cellcolor{cellblue!80}\best{74.31} & \cellcolor{cellblue!80}\best{72.22} & \cellcolor{cellblue!60}\second{80.80} & \cellcolor{cellblue!80}\best{75.63} & \cellcolor{cellblue!77}69.11 & \cellcolor{cellblue!77}\second{76.35} \\
\bottomrule
\end{tabular}}
\end{table}


\subsection{Which augmentations are useful?}
We next ask which augmentations create an informative teacher--student asymmetry that can serve as an effective predictive learning signal.
Each configuration below is a complete training run that differs only in $T$; the training data, optimizer, step budget, and evaluation protocol are held fixed. Due to limited compute, these analysis runs are evaluated with greedy decoding at a $2{,}048$-token budget; their absolute numbers are therefore internally comparable but sit slightly below the $4{,}096$-token rows of Table~\ref{tab:main}.

\begin{figure}[h]
\centering
\includegraphics[width=\linewidth]{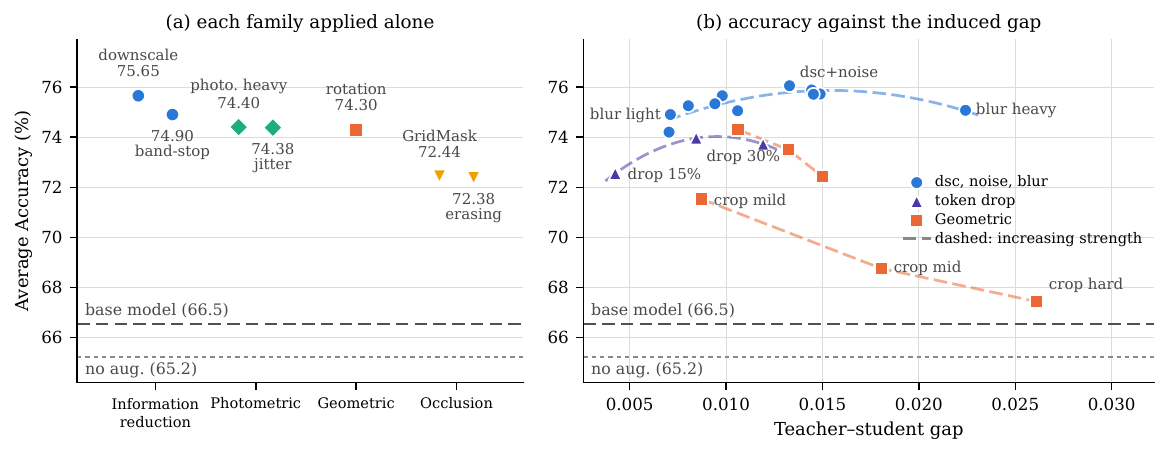}
\caption{\textbf{(a)} Performance of each augmentation family applied alone;
\textbf{(b)} Accuracy vs.\ the teacher--student predictive gap induced by the augmentation. We measure the teacher-student gap as the token-level JS Divergence
between the teacher and the student distributions, averaged over the first ten steps of
training. }
\label{fig:families}
\label{fig:induced_gap}
\end{figure}

\textbf{All augmentation families are effective.}
\label{sec:exp:families}\label{sec:exp:attr}
Under this fixed analysis protocol, the base model achieves $70.58\%$. Symmetric self-distillation with $T$ equal to the identity reaches only $65.21\%$, slightly degrading the initial model. In contrast, applying each augmentation family independently yields $75.65\%$ for information reduction, $74.40\%$ for photometric transformations, $74.30\%$ for geometric transformations, and $72.44\%$ for occlusion (Figure~\ref{fig:families}(a)). Thus, every family substantially improves over both the base model and unaugmented self-distillation.

\textbf{Strength matters: neither too weak nor too strong.}
\label{sec:exp:scale}\label{sec:exp:composition}
Across all augmentation families, performance first improves and then declines as the strength increases. We show this in Figure~\ref{fig:induced_gap}(b) by plotting the averaged performance against the teacher-student gap induced by the augmentation.
We measure teacher-student gap as the token-level JS Divergence between the teacher and the student distributions, averaged over the first ten steps of training, which aligns well with the strength parameters of each augmentation (e.g., heavy blur induces a larger teacher--student gap than light blur).
Resolution reduction peaks at $75.65\%$ for a scale range of $0.3$--$0.6$, compared with $75.25\%$ and $75.05\%$ at weaker and stronger strength. Blur follows the same trend, rising from $74.20\%$ to $75.71\%$ before falling back to $75.07\%$, as does visual-token dropping, which moves from $72.55\%$ up to $73.96\%$ and back down to $73.72\%$. More generally, plotting accuracy against the induced teacher--student predictive gap reveals a consistent pattern: performance increases with the gap up to a JS divergence of roughly $0.014$, then declines beyond it. Thus, useful asymmetry requires sufficient, but not excessive, perturbation.

\textbf{A larger gap is not a better gap: the augmentation must preserve question-relevant information.}
\label{sec:exp:crop}\label{sec:exp:companions}
Cropping exposes this limitation. Unlike information-reduction augmentations, its performance decreases monotonically with strength, falling from $71.53\%$ to $68.76\%$ and then $67.44\%$ across mild, moderate, and strong crops (Figure~\ref{fig:induced_gap}(b)). The strongest crop produces the largest predictive gap in our study yet achieves only $67.44\%$, barely above unaugmented self-distillation. Even the mildest setting, whose induced gap coincides with that of the best resolution and blur configurations, trails them by more than $3\%$. The degradation is especially pronounced on benchmarks whose answers depend on localized or global visual evidence: moving from moderate to strong cropping reduces V*Bench by $2.1\%$, HR-Bench~8K by $2.4\%$, and MME-RealWorld by $2.1\%$. Cropping may remove the evidence required to answer the question altogether. The resulting discrepancy is therefore large partly because the student's input has become unanswerable, and such a gap provides little useful learning signal.


Overall, useful asymmetry is governed by both magnitude and semantics. The augmentation must create a sufficient information gap to induce learning, while the structure of that gap determines which perceptual capabilities are emphasized. Downscaling with Gaussian noise offers the most robust default across benchmarks.

\subsection{Ablation and analysis}
\label{sec:exp:objective}

We ablate the effectiveness of augmentation and EMA teacher in Table~\ref{tab:component}.

\begin{table}[t]
\centering
\small
\setlength{\tabcolsep}{4pt}
\caption{Component ablation on \method{}-4B. \emph{w/o aug} removes the view asymmetry by showing the student the same clean image as the teacher; \emph{w/o EMA} keeps the augmentation but freezes the teacher at the base model.}
\label{tab:component}
\begin{tabular}{lccccccc}
\toprule
\textbf{Method} & \textbf{V*} & \textbf{Zoom} & \textbf{HR-4K} & \textbf{HR-8K} & \textbf{MME-RW} & \textbf{MME-RW-CN} & \textbf{\avg{}} \\
\midrule
\method{} & \cellcolor{cellblue!72}\best{87.43} & \cellcolor{cellblue!80}\best{57.99} & \cellcolor{cellblue!8}84.88 & \cellcolor{cellblue!80}\best{83.62} & \cellcolor{cellblue!80}\best{72.87} & \cellcolor{cellblue!80}\best{71.29} & \cellcolor{cellblue!80}\best{76.35} \\
w/o Aug. & \cellcolor{cellblue!8}79.06 & \cellcolor{cellblue!8}50.18 & \cellcolor{cellblue!70}\best{86.50} & \cellcolor{cellblue!8}80.62 & \cellcolor{cellblue!8}62.85 & \cellcolor{cellblue!8}63.90 & \cellcolor{cellblue!8}70.52 \\
w/o EMA (fixed teacher) & \cellcolor{cellblue!80}\best{88.48} & \cellcolor{cellblue!44}54.08 & \cellcolor{cellblue!80}86.75 & \cellcolor{cellblue!71}83.25 & \cellcolor{cellblue!75}72.15 & \cellcolor{cellblue!77}70.98 & \cellcolor{cellblue!75}75.95 \\
\bottomrule
\end{tabular}
\end{table}

\textbf{Asymmetry is necessary.} Removing the student-view augmentation while keeping the objective (\emph{w/o Aug.}, Table~\ref{tab:component}) collapses the gain back to the base level: $70.52\%$ against the base model's $70.68\%$ on the six perception benchmarks. With no information gap, the objective amplifies the teacher's confident errors instead of correcting the student's. 

\textbf{EMA teacher.} Freezing the teacher at the base model (\emph{w/o EMA}) costs only $0.40\%$, still reaching $75.95\%$: a teacher that never improves still recovers $93\%$ of the gain. The supervision therefore comes more from the constructed asymmetry than from teacher self-improvement.

We analyze the two choices that instantiate the objective: the teacher update rate and the divergence minimized by the student.

\textbf{\method{} is robust to the teacher update rate.} In Table~\ref{tab:ema}, across decay values from $0.95$ to $0.999$, the six-benchmark average remains within $0.8\%$ and shows no monotonic trend, indicating that \method{} is not sensitive to the precise teacher update rate. This robustness suggests that the teacher's gradual evolution is not the main driver of performance. Rather, the effective learning signal comes from the teacher--student view asymmetry: the teacher provides a less-degraded reference distribution, while the student is trained to recover from the degraded view. Consistent with this interpretation, even the frozen-teacher variant in Table~\ref{tab:component} reaches $75.95\%$, only $0.40\%$ below the default. Thus, EMA mainly serves as a stable implementation choice, whereas the essential supervision comes from the constructed asymmetry.

\textbf{\method{} benefits from a balanced divergence.} Table~\ref{tab:alpha} analyzes the divergence hyperparameter by comparing the generalized Jensen--Shannon divergence with forward-KL and reverse-KL. Both are consistently
worse than JSD (the symmetric midpoint) on each reported benchmark and on average, with the ordering
JSD $>$ reverse KL $>$ forward KL. This pattern indicates that asymmetric-view distillation requires a divergence that balances coverage and selectivity. Forward KL is too coverage-seeking: it forces the student to match probability mass assigned by the teacher, including fine-grained details that may be unavailable in the degraded student view. Reverse KL is too mode-seeking: it lets the student concentrate on high-confidence modes and discard softer corrective signals. JSD provides a middle ground, transferring useful teacher information while avoiding excessive pressure to imitate inaccessible visual details.

\begin{table}[t]
\begin{minipage}[t]{0.48\linewidth}
\centering
\small
\setlength{\tabcolsep}{3pt}
\caption{Comparison of performance under different teacher update rates (EMA). \avg{} is over 6 perception benchmarks.}
\label{tab:ema}
\resizebox{\linewidth}{!}{%
\begin{tabular}{lccc}
\toprule
\textbf{Teacher decay} & \textbf{MME-RW} & \textbf{MME-RW-CN} & \textbf{\avg{}} \\
\midrule
$0.95$ & \cellcolor{cellblue!80}\best{72.87} & \cellcolor{cellblue!46}71.29 & \cellcolor{cellblue!80}\best{76.35} \\
$0.99$ & \cellcolor{cellblue!46}72.53 & \cellcolor{cellblue!19}71.07 & \cellcolor{cellblue!8}75.56 \\
$0.999$ & \cellcolor{cellblue!22}72.29 & \cellcolor{cellblue!80}\best{71.56} & \cellcolor{cellblue!48}76.00 \\
\bottomrule
\end{tabular}}
\end{minipage}\hfill
\begin{minipage}[t]{0.48\linewidth}
\centering
\small
\setlength{\tabcolsep}{3pt}
\caption{Comparison of different divergences. \avg{} is over 6 perception benchmarks.}
\label{tab:alpha}
\resizebox{\linewidth}{!}{%
\begin{tabular}{lccc}
\toprule
$\boldsymbol{\alpha}$ & \textbf{MME-RW} & \textbf{MME-RW-CN} & \textbf{\avg{}} \\
\midrule
$0.0$ (forward KL) & \cellcolor{cellblue!8}71.92 & \cellcolor{cellblue!8}69.75 & \cellcolor{cellblue!8}74.74 \\
$0.5$ (JSD) & \cellcolor{cellblue!80}\best{74.15} & \cellcolor{cellblue!80}\best{71.34} & \cellcolor{cellblue!80}\best{76.05} \\
$1.0$ (reverse KL) & \cellcolor{cellblue!24}72.43 & \cellcolor{cellblue!65}71.00 & \cellcolor{cellblue!49}75.49 \\
\bottomrule
\end{tabular}}
\end{minipage}
\end{table}



\section{Conclusion}
\label{sec:conclusion}

\method{} proposes a novel method to creates informative teacher--student asymmetry without privileged annotations by subtracting task-relevant information from the student view. The teacher observes the original image, while the student learns from a degraded view; this information gap turns the model's own predictions into a perception-aligned training signal. Further analysis shows that the effectiveness of \method{} is governed by the magnitude and semantics of the induced information gap. Augmentation must remove enough task-relevant evidence to create a useful learning signal, but not so much that it changes the underlying question or makes recovery impossible. 
Across benchmarks and model scales, \method{} improves both fine-grained perception and mathematical reasoning, distinguishing \method{} from privileged-supervision methods that improve perception but can weaken reasoning, and from self-rewarding RL methods that preserve reasoning but provide weaker perceptual gains.


\bibliographystyle{plainnat}
\bibliography{references}

\end{document}